\documentclass[conference]{IEEEtran}
\IEEEoverridecommandlockouts
\usepackage{cite}
\usepackage{amsmath,amssymb,amsfonts}
\usepackage{algorithmic}
\usepackage{graphicx}
\usepackage{textcomp}
\usepackage{xcolor}
\def\BibTeX{{\rm B\kern-.05em{\sc i\kern-.025em b}\kern-.08em
    T\kern-.1667em\lower.7ex\hbox{E}\kern-.125emX}}
\begin{document}

\title{I am Only Happy When There is Light: The Impact of Environmental Changes on Affective Facial Expressions Recognition
}

\author{\IEEEauthorblockN{Doreen Jirak, Alessandra Sciutti}
\IEEEauthorblockA{\textit{Contact Unit} \\
\textit{Italian Institute of Technology}\\
Genoa, Italy \\
\{doreen.jirak, alessandra.sciutti\}@iit.it}
\and
\IEEEauthorblockN{Pablo Barros}
\IEEEauthorblockA{\textit{Sens.AI} \\
\textit{R\&D Center - Sony}\\
Brussels, Belgium \\
pablo.barros@sony.com}
\and
\IEEEauthorblockN{Francesco Rea}
\IEEEauthorblockA{\textit{RBCS} \\
\textit{Italian Institute of Technology}\\
Genoa, Italy \\
francesco.rea@iit.it}

}


\maketitle

\begin{abstract}
Human-robot interaction (HRI) benefits greatly from advances in the machine learning field as it allows researchers to employ high-performance models for perceptual tasks like detection and recognition. Especially deep learning models, either pre-trained for feature extraction or used for classification, are now established methods to characterize human behaviors in HRI scenarios and to have social robots that understand better those behaviors. As HRI experiments are usually small-scale and constrained to particular lab environments, the questions are how well can deep learning models generalize to specific interaction scenarios, and further, how good is their robustness towards environmental changes? These questions are important to address if the HRI field wishes to put social robotic companions into real environments acting consistently, i.e. changing lighting conditions or moving people should still produce the same recognition results. In this paper, we study the impact of different image conditions on the recognition of arousal and valence from human facial expressions using the FaceChannel framework \cite{Barro20}. Our results show how the interpretation of human affective states can differ greatly in either the positive or negative direction even when changing only slightly the image properties. We conclude the paper with important points to consider when employing deep learning models to ensure sound interpretation of HRI experiments.    
\end{abstract}


\section{Introduction}
Facial Expression Recognition (FER) emerged as an important topic in affective computing with applications for social robots which are enabled to show and infer emotional behaviors (for recent reviews on the topics we would like to refer the interested reader to \cite{Spezi20, Stock21}). Recent advances in machine learning, especially deep learning, facilitate HRI research as popular models like ResNet or VGG16 can be used, e.g., as state-of-the-art models for object recognition to provide robust features and classification abilities to a robot in HRI scenarios. Similarly, it is desirable to have a FER framework to be used ``out-of-the-box" to study affective states and emotional behaviors in humans for collaborative tasks in domestic or working environments. \cite{Barro20} introduced the so-called FaceChannel, a deep learning framework based on convolutional neural networks (CNN) which implements both affective state recognition in terms of arousal and valence, as well as classic categorical emotion classification \cite{Baltr18}. Other frameworks like the OpenFace consider so-called action units that represent facial muscle movement involved in emotional expressions \cite{Ekman70}. However, all deep learning models are data-driven, thus they might not well represent particular HRI scenarios and may not capture differences in the environment such as lighting conditions as is the case for, e.g., autonomously moving robots. Therefore, in this paper, we address the question of robustness towards environmental changes for the recognition performance of arousal and valence. We use the FaceChannel framework as it provides easy access to FER and its lightweight implementation allows the recognition in real-time, which renders this software interesting for robotic applications.

As we want to understand if and how researchers address the robustness of FER with state-of-the-art deep learning architectures, we chose publications that deal with ``in-the-wild" datasets or environments as they refer to the reliability of FER classifiers. \cite{Ruiz18, Webb20} introduce a learning model based on stacked autoencoders and CNNs, called SCAE, that can learn facial expressions invariant to illumination changes and face poses. The authors show that pre-training a CNN with their method achieves up to 28\% performance boost on a test set obtained from unconstrained environments compared to a standard CNN. The authors highlight the comparative performance of their model compared to state-of-the-art methods applied on popular datasets like JAFFE and CK+ and also the significance of heterogeneous datasets when collected by the humanoid NAO robot for robust recognition. 
\cite{Han21} present a study using a mobile robot platform and achieved good performance with an ensemble method for ``in-the-wild" images. However, it is a small-scale study because the authors used only 60 images for testing.
\cite{Arnau22} address the influence of what the authors call "exogenous" variables like rotations and scale that can be considered image manipulation on the recognition of digital numbers and shapes and introduce the THIN architecture (THrowable Information Networks), which the authors also apply to FER. Although they could show the superior performance of their model compared to state-of-the-art, the authors target ``identity" and do not opt for image manipulations of the faces.
\cite{Ramis20} employed a social robot that motivates participants to perform certain facial expressions. The goal of this study is to collect data for comparison with a CNN-based recognition framework and experts. The authors address the importance of dataset diversity, referred to as "cross dataset" training of CNNs but the authors did not further integrate this approach into their framework.

\section{Methodology}
\begin{figure}[htbp]
\centerline{\includegraphics[width=0.5\textwidth]{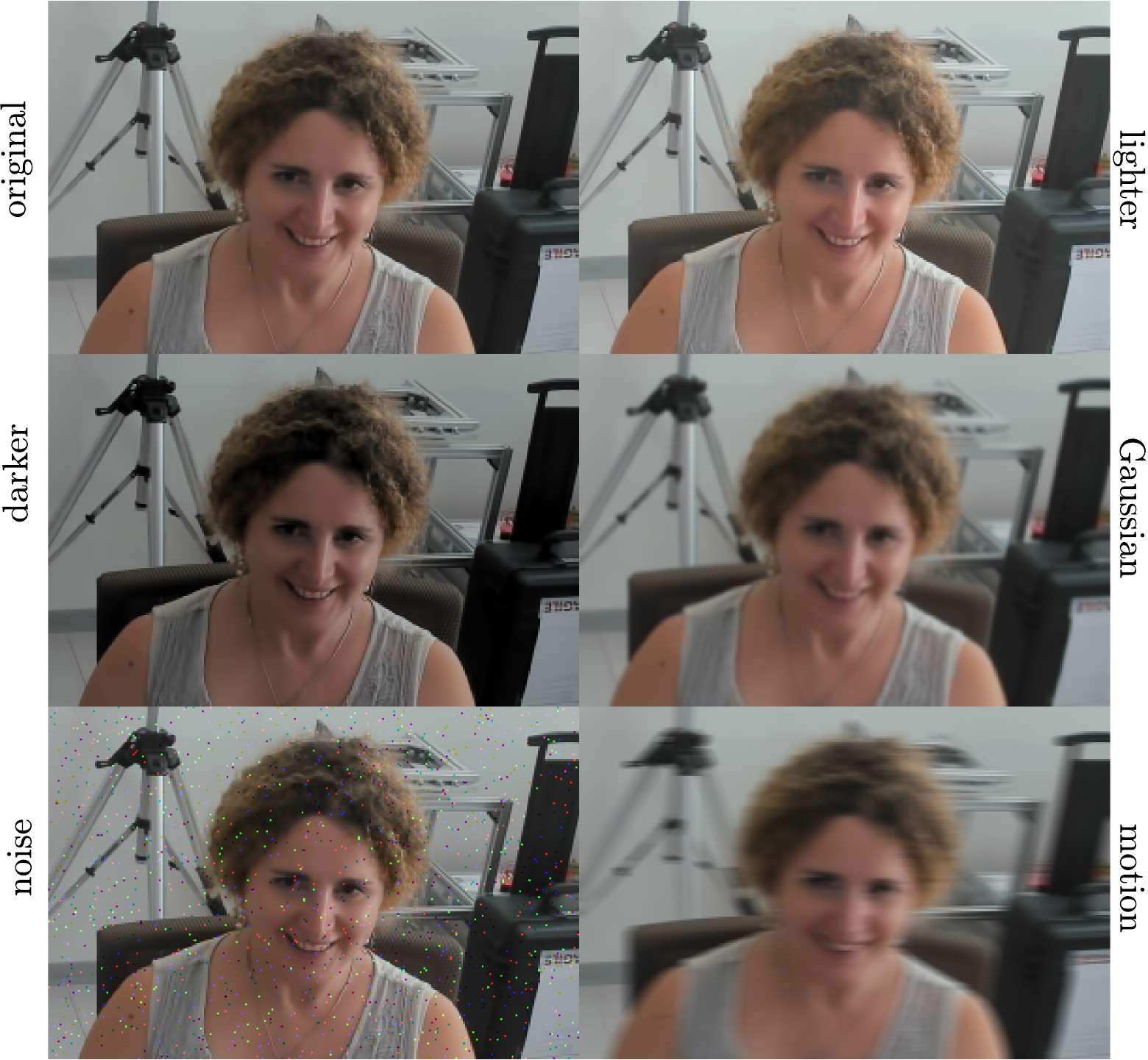}}
\caption{Example image of a participant (original) and the image conditions that we apply to change the image properties. Note that for all conditions the, e.g., smiling of the person is visible to ensure fair comparisons of the FaceChannel output between the original and image conditions.}
\label{fig:sub16}
\end{figure}

We use the FaceChannel framework for affective state recognition, i.e., arousal and valence. Arousal is a measure of excitement while valence describes the level of (un)pleasantness. Both metrics are evaluated continuously in the range $[-1;1]$, i.e. depending on the value pair, a human can be described as positively excited (both values close to 1) or sad (low arousal and negative valence), etc. The FaceChannel used to obtain arousal-valence is trained on the AffectNet dataset \cite{Molla17} using fewer parameters than other models like the mentioned VGG16 but with similar to superior performance. Due to its lightweight implementation, the FaceChannel can be integrated into robotic systems or HRI scenarios. 
As a baseline for the arousal-valence recognition, we use a subset of a dataset recorded at the Istituto Italiano di Tecnologia (IIT) for a study addressing human performance under cognitive load \cite{Aoki22}. The 14 participants solved different tasks sequentially in presence of the humanoid robot iCub, which showed either so-called ``non-social" or ``social" behaviors. As a baseline, the participants performed the task without the presence of the robot. The analysis of the ``social" condition revealed mostly positive arousal-valence patterns elicited by the participants compared to the mixed results in the ``non-social" condition or mostly neutral states in the baseline \cite{Jirak22}. Hence, we consider the ``social" subset as reasonable data baseline. First, we convert the original video sequences into images (frames) and delete all images from the experiment instructions as the participants looked mostly sideways or wore a mask due to Covid19 restrictions. Thus, we obtain around $\approx$ 6000-7000 images per participant. Then, we alter the original images like this: We create a bounding box for each participant that contains only the face and crop all images. Then, we alter the brightness so that the images appear lighter or darker. Also, we introduce image blur by applying a 2D Gaussian filter (standard deviation $\sigma$=1). Additionally, we apply so-called salt-and-pepper noise to the images with a level of 0.01, i.e., there is a 1\% chance for a pixel to be flipped. Finally, we simulate the horizontal camera motion of 10 pixels. Figure \ref{fig:sub16} shows exemplary the resultant images. It also shows that the facial expression is still perceivable, i.e., the image is not too dark or distorted. 

\section{Results and Evaluation}
\begin{figure}[htbp]
\centerline{\includegraphics[width=0.5\textwidth]{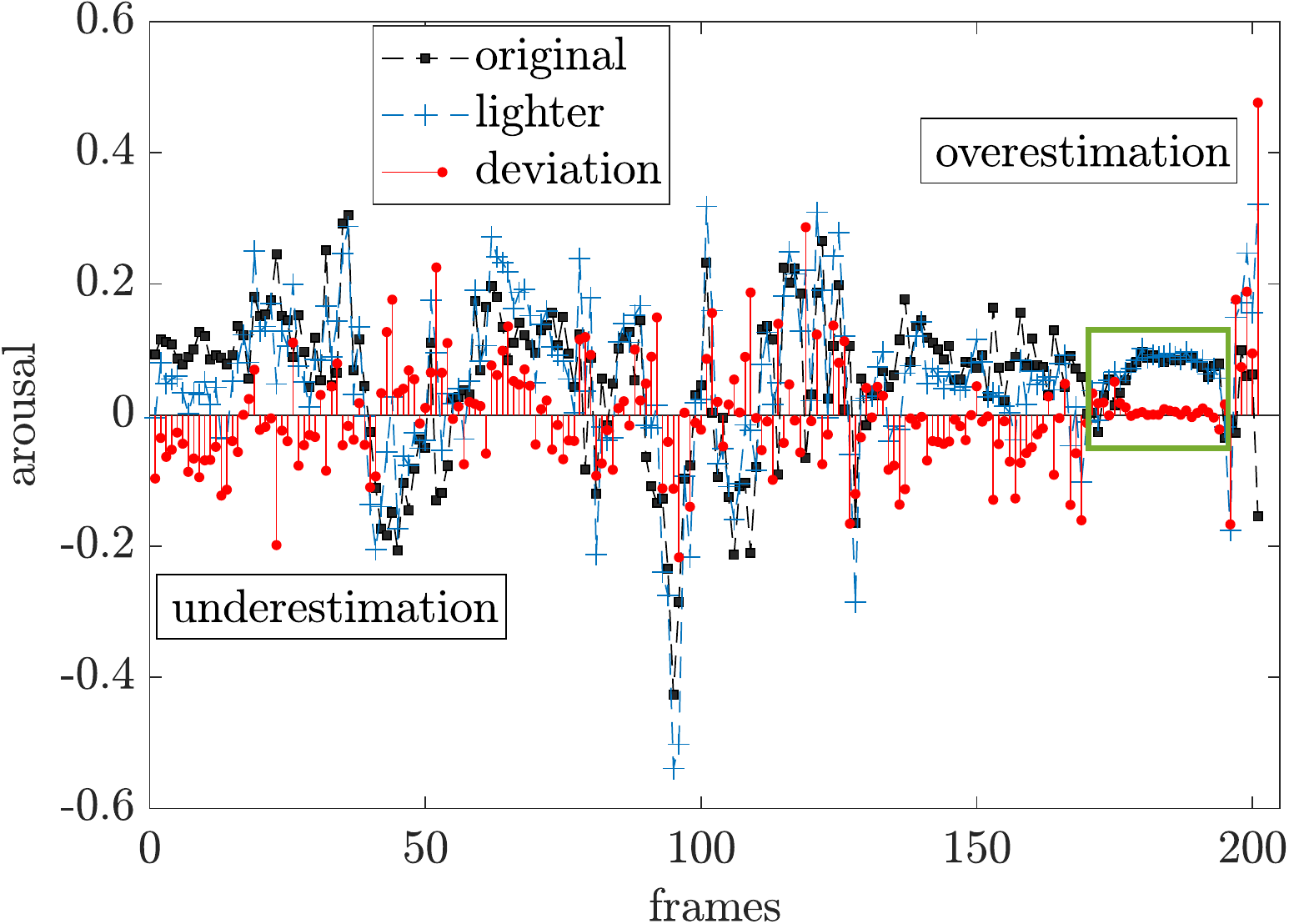}}
\caption{Exemplary sequence of an original arousal sequence (black) compared to the resultant arousal sequence obtained from images with lighter brightness (blue). The red bars denote the amount of deviations in either positive or negative direction. The green box shows a segment where the FaceChannel produces a good agreement between the arousal values.}
\label{fig:sub16_deviation_a}
\end{figure}

\begin{figure}[htbp]
\centerline{\includegraphics[width=0.5\textwidth]{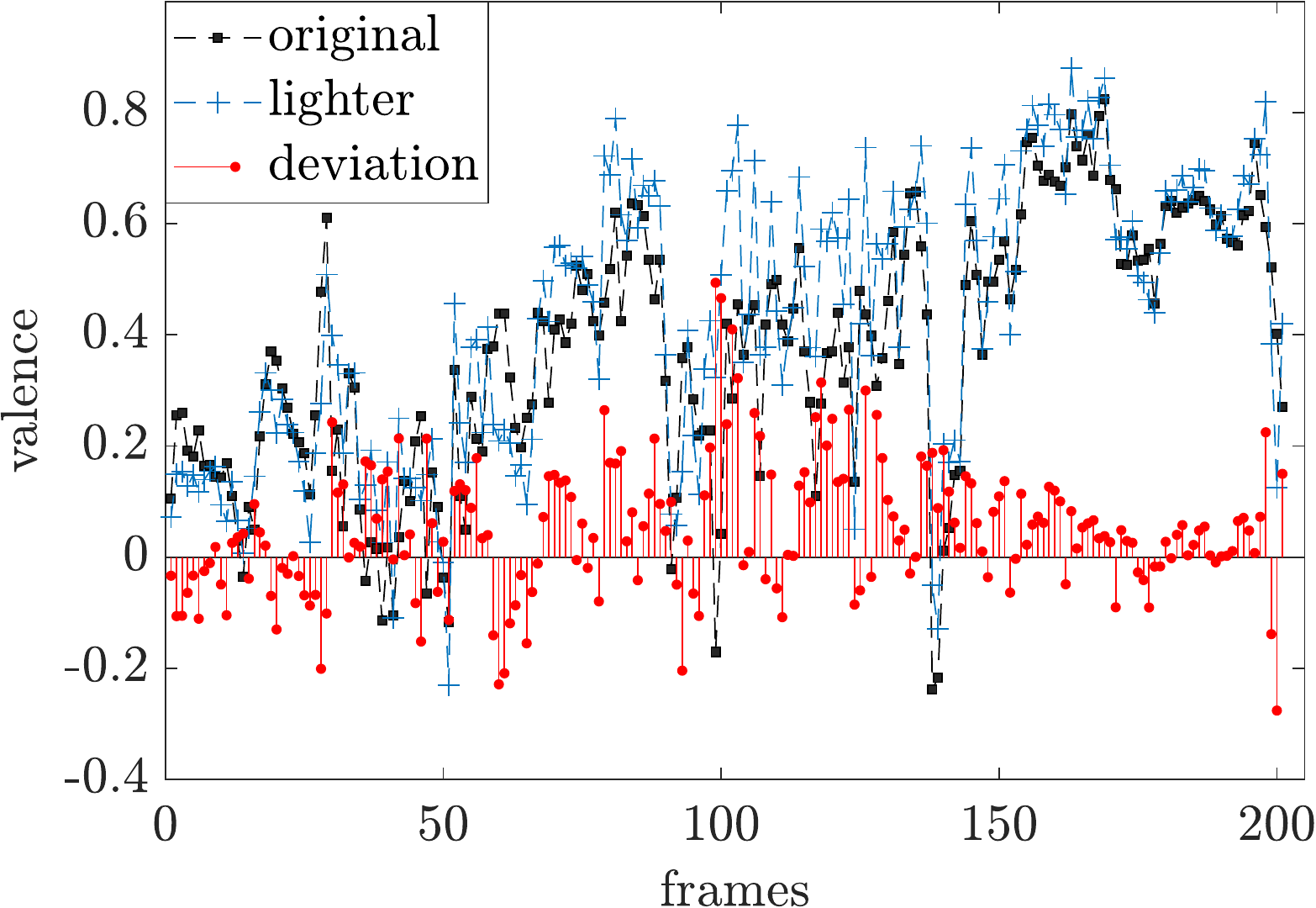}}
\caption{Exemplary sequence of an original valence sequence (black) compared to the resultant valence sequence obtained from images with lighter brightness (blue). The red bars denote the amount of deviations in either positive or negative direction.}
\label{fig:sub16_deviation_v}
\end{figure}

We present our results both from the qualitative and the quantitative point where the baselines for comparison are the arousal-valence sequences obtained from the FaceChannel applied to the original data described above. We first use a scatter plot to visualize the data distribution of all conditions (left out for brevity). For the `lighter' and `darker' conditions, we observe mostly slight deviations from the original arousal-valence distribution with some peaks in the negative or positive direction. The `Gaussian' and `noise' conditions reveal a large spread over the arousal-valence dimension, which distorts the original arousal-valence responses. Interestingly, the `motion' condition seems to contract the arousal-valence dimension as we observe clusters around the neutral state (around (0,0)). We then compare the time series for arousal and valence individually to analyze the trend of the observed difference, i.e., we want to see the trend of dispersion. We simply calculate, e.g. for arousal, $a_{cond}-a_{orig}$, which is either positive (overestimation) or negative (underestimation) and quantifies the distance between the original data ($a_{orig}$) and the respective image condition ($a_{cond}$). Figures \ref{fig:sub16_deviation_a} and \ref{fig:sub16_deviation_v} show a result of this simple scheme. We see how the `lighter' condition deviates in the positive and negative direction mostly mildly but with some peaks diverging significantly from the original value (cf. Figure \ref{fig:sub16_deviation_a}). However, we also see segments that show consistent classification between the original data and the `lighter' condition, shown in the green box. However, the figures are only exemplary, and our analysis reveals much larger deviations for the `Gaussian`, `noise', and `motion' conditions. For a better understanding of the trends, we then computed the frequency of positive and negative values, i.e., how often they are represented in a sequence, and calculated the percentage over all participants and per image condition. 
The `noise' and `darker' conditions distort the original arousal values into the positive direction, while a clear trend toward negative values becomes visible for the `Gaussian' and `motion' conditions. This result supports the observed ``attenuation' effect of the arousal dimension. The opposite is the case for valence, where the latter conditions show trends towards positive values. Both conditions seem to distract the interpretation of arousal-valence towards a more calm and pleasant affective state and, thus, may underestimate the human's real emotion. The `darker' and `noise' conditions impact the valence more negatively, i.e., both conditions can distort the interpretation of arousal-valence towards a negatively excited affective state, hence, overestimating the affective state of a calm person. Finally, the `lighter' condition has a very mixed result on the positive and negative trend but with fewer deviations from the original value, hence this condition seems to deliver the most reliable results.

\begin{figure}[htbp]
\centerline{\includegraphics[width=0.5\textwidth]{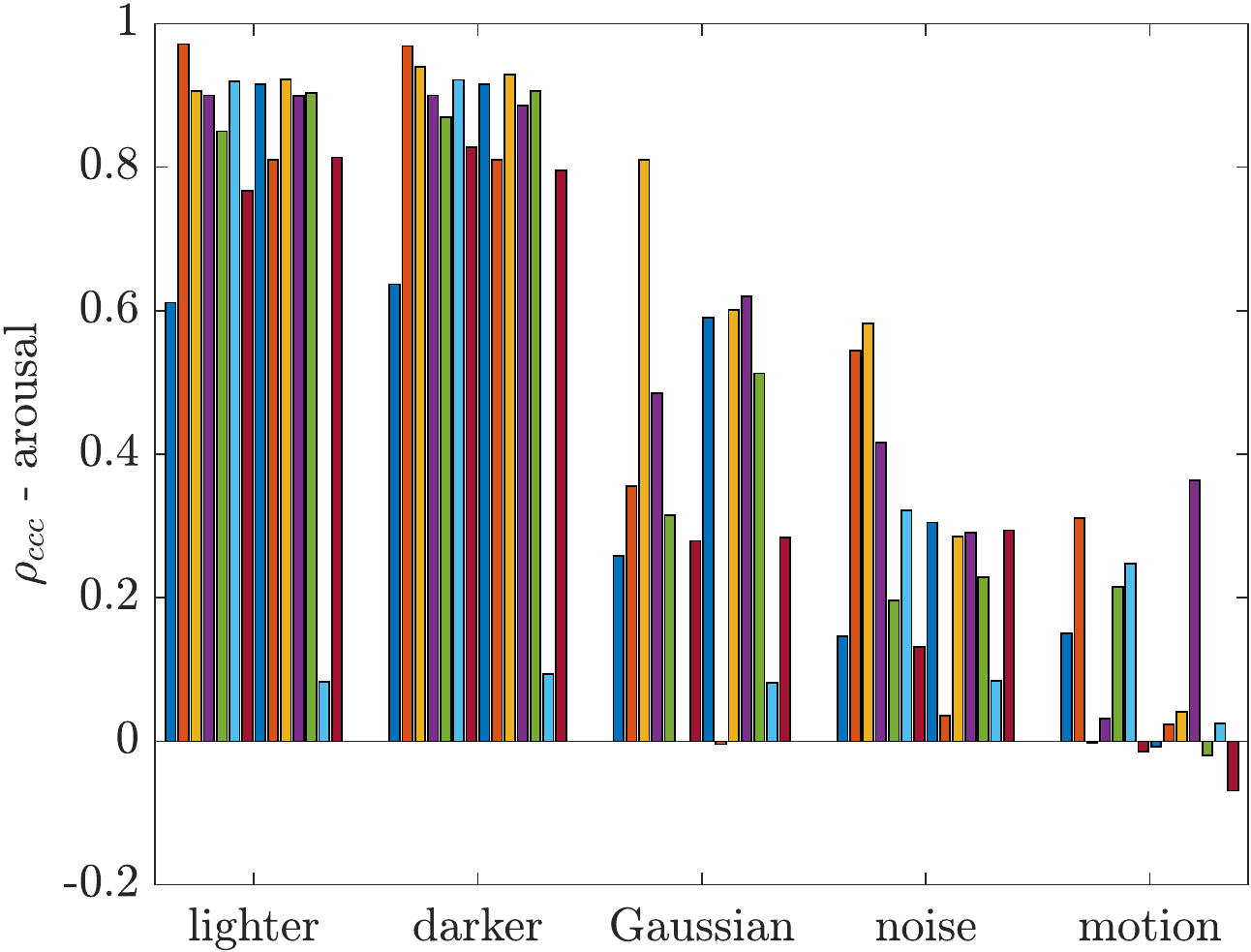}}
\caption{Distribution of the concordance correlation coefficient  $\rho_{ccc}\in [-1;1]$ between the baseline of the arousal compared to the image condition over all 14 participants. While the `lighter' and `darker' arousal sequences show high agreement with the original data, the distributions for the other conditions show a clear decline. }
\label{fig:ccc_a}
\end{figure}

\begin{figure}[htbp]
\centerline{\includegraphics[width=0.5\textwidth]{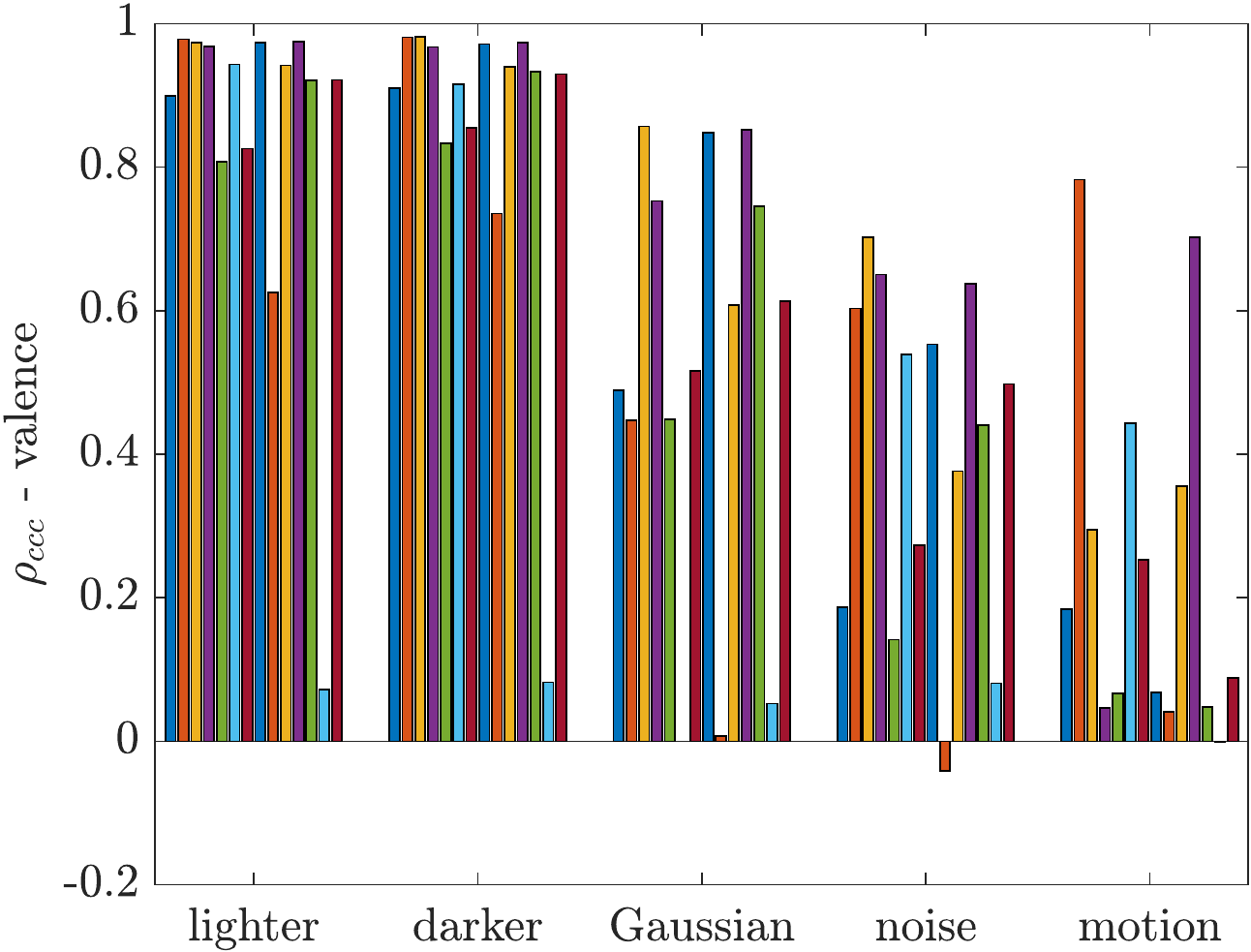}}
\caption{Distribution of the concordance correlation coefficient $\rho_{ccc}\in [-1;1]$ between the baseline of the valence compared to the image condition over all 14 participants. Similarly to the arousal distribution, we observe a decline in the correlation values for the `noise' and `motion` condition but less for the `Gaussian' condition.}
\label{fig:ccc_v}
\end{figure}

We compute the concordance correlation coefficient ($\rho_{ccc}$, cf. \ref{eq:ccc}) which is measure of agreement between two data sequences $x$ and $y$. We use this coefficient to quantify the described deviations. It is defined as:
\begin{equation}
\rho_{ccc}=\frac{2\rho_{p}\sigma_x\sigma_y}{\sigma_{x}^2+\sigma_{y}^2+(\mu_x-\mu_y)^2}
\label{eq:ccc}
\end{equation}
where $\rho_p$ is the Pearson correlation coefficient, $\sigma^2$ the variance of the two variables $x$ and $y$ (here the original arousal-valence sequence and sequences per image condition), and $\mu$ the mean values, respectively. Figures \ref{fig:ccc_a} and \ref{fig:ccc_v} display the distribution of $\rho_{ccc}$ over all participants and image condition for both arousal and valence. The results support our qualitative impression from the visualizations (scatter plot, trends in sequences). While the `lighter' and `darker' conditions show a high correlation with the original data, which means that the conditions have only a small impact on the arousal-valence evaluations, we observe a clear decline for the remaining three conditions. Especially the `noise' and the `motion' conditions reveal only small to even no correlation, where the arousal dimension seems more affected than the valence dimension. For a few individual cases, we observe even slight negative correlations. The range of $\rho_{ccc}$ is further outlined in Table \ref{tab:ccc}, which supports that the arousal dimension is more affected by the `Gaussian', `noise', and `motion' condition compared to the valence dimension.

\begin{table}[]
\centering
\caption{CCC - Arousal \& Valence}
\label{tab:ccc}
\begin{tabular}{|c|cc|cc|}
\hline
\multicolumn{1}{|l|}{} & \multicolumn{2}{c|}{Arousal} & \multicolumn{2}{c|}{Valence} \\ \hline
condition & \multicolumn{1}{c|}{min $\rho_{ccc}$} & max $\rho_{ccc}$ & \multicolumn{1}{c|}{min $\rho_{ccc}$} & max $\rho_{ccc}$ \\ \hline
lighter & \multicolumn{1}{c|}{0.0825} & 0.9715 & \multicolumn{1}{c|}{0.0725} & 0.9784 \\ \hline
darker & \multicolumn{1}{c|}{0.0941} & 0.9690 & \multicolumn{1}{c|}{0.0823} & 0.9818 \\ \hline
Gaussian & \multicolumn{1}{c|}{-0.0040} & 0.8099 & \multicolumn{1}{c|}{0.0071} & 0.8568 \\ \hline
noise & \multicolumn{1}{c|}{0.0358} & 0.5824 & \multicolumn{1}{c|}{-0.0414} & 0.7022 \\ \hline
motion & \multicolumn{1}{c|}{-0.0690} & 0.3639 & \multicolumn{1}{c|}{-0.0015} & 0.7825 \\ \hline
\end{tabular}
\end{table}

\section{Discussion and Conclusion}
Our study shows how FER frameworks like the FaceChannel \cite{Barro20} can limit their application when introducing small image modifications. In particular, our study on arousal-valence sequences revealed high distortions in the arousal-valence dimension when simulating camera movements as is the case for the `Gaussian' blur (e.g. camera focus adjustment) and real movement (`motion') even for a small amount of pixels involved. Also, darker light environments and possible noise (e.g. image transmission, artifacts) impact arousal-valence recognition. Consequently, minor changes in the environmental condition can have a huge impact on the affective state's interpretation of participants in HRI studies. Our analysis revealed that, e.g., participants with mostly positive arousal-valences responses can be misinterpreted by a robot as calm or even bored people because camera motion attenuates the corresponding signals. On the other hand, noise and different lighting conditions tend to overestimate the real affective states as shown by deviations into the positive direction. Although the issues might be known to many researchers in the field, during our literature search we mostly encountered tailored learning solutions in the field of FER, which complicates reproducibility, or pure application of established software on rather small datasets without any variations. ``In-the-wild" solutions are a step in the right direction but more studies are needed, especially as FER is already subject to misinterpretations due to insufficient emotion categorizations and neglect of context and other modalities. Therefore, we would like to point out the most critical aspects to address in HRI to enforce robustness in learning models for robotic applications beyond strict lab conditions:
\begin{itemize}
    \item Use data augmentation: many libraries such as PyTorch\footnote{\url{https://pytorch.org/}} allow data augmentation during training, i.e. the image manipulations as carried out in this study can be easily integrated into the learning process. It enhances the usually small data corpora collected in HRI studies and avoids time-consuming labeling. 
    \item Re-train the last layer: recognition models usually rely on CNNs, whose last layer can be re-trained to integrate the image properties of a particular dataset. A recent paper provides a study on this topic \cite{Barro22}.
    \item Go multi-modal: the interpretation of FER can be highly subjective. Therefore, it is common to add physiological sensors to measure electro-dermal activity or to measure the heart rate. However, the instruments need individual calibrations and are highly error-prone to a participant's movement. A holistic vision approach can corporate other modalities such as eye gaze or (body) gestures to give also contextual information. For instance, we observed participants who cover their face (hands-over-face gesture) or shrug their shoulders, which can be signs of unpleasantness or negative surprise.
    \item Make careful claims: a deep learning model may offer great support for HRI studies but as we have shown, little modifications in the image may render an excited person into a bored one.
\end{itemize}

We are also aware that our study has some limitations. First, 14 participants can be considered a small number but due to the lack of similar studies, we opt for establishing a reasonable data analysis pipeline first. For future work, we would like to study other datasets and also consider emotion classification using different models. It could be interesting to analyze their deviations in labeling and the impact on human behavior interpretation. Also, more sophisticated analysis tools or new metrics for robustness could be established in the future. Finally, other image conditions might be added such as small rotations or face occlusions. For those cases, it is also an important question of how to interpolate affective states sensitively, e.g. when persons move in the scene and do not always show their face. We hope our current work inspires other researchers in the field of deep learning and affective state recognition to develop and analyze FER models for effective and sound employment in HRI research.

\section*{Acknowledgment}
This work has been supported by the project VOJEXT under the European Union's Horizon 2020 innovation programme, G.A. No 952197.

\bibliographystyle{IEEEtran}
\bibliography{SCIAR22_JirakSciuttiBarrosRea}

\begin{thebibliography}{10}
\providecommand{\url}[1]{#1}
\csname url@samestyle\endcsname
\providecommand{\newblock}{\relax}
\providecommand{\bibinfo}[2]{#2}
\providecommand{\BIBentrySTDinterwordspacing}{\spaceskip=0pt\relax}
\providecommand{\BIBentryALTinterwordstretchfactor}{4}
\providecommand{\BIBentryALTinterwordspacing}{\spaceskip=\fontdimen2\font plus
\BIBentryALTinterwordstretchfactor\fontdimen3\font minus
  \fontdimen4\font\relax}
\providecommand{\BIBforeignlanguage}[2]{{%
\expandafter\ifx\csname l@#1\endcsname\relax
\typeout{** WARNING: IEEEtran.bst: No hyphenation pattern has been}%
\typeout{** loaded for the language `#1'. Using the pattern for}%
\typeout{** the default language instead.}%
\else
\language=\csname l@#1\endcsname
\fi
#2}}
\providecommand{\BIBdecl}{\relax}
\BIBdecl

\bibitem{Barro20}
P.~Barros, N.~Churamani, and A.~Sciutti, ``The facechannel: A light-weight deep
  neural network for facial expression recognition.'' in \emph{2020 15th IEEE
  International Conference on Automatic Face and Gesture Recognition (FG
  2020)}, 2020, pp. 652--656.

\bibitem{Spezi20}
\BIBentryALTinterwordspacing
M.~Spezialetti, G.~Placidi, and S.~Rossi, ``Emotion recognition for human-robot
  interaction: Recent advances and future perspectives,'' \emph{Frontiers in
  Robotics and AI}, vol.~7, 2020. [Online]. Available:
  \url{https://www.frontiersin.org/articles/10.3389/frobt.2020.532279}
\BIBentrySTDinterwordspacing

\bibitem{Stock21}
R.~Stock-Homburg, ``Survey of emotions in human--robot interactions:
  Perspectives from robotic psychology on 20 years of research,''
  \emph{International Journal of Social Robotics}, pp. 1--23, 2021.

\bibitem{Baltr18}
T.~Baltrusaitis, A.~Zadeh, Y.~C. Lim, and L.-P. Morency, ``Openface 2.0: Facial
  behavior analysis toolkit,'' in \emph{2018 13th IEEE International Conference
  on Automatic Face Gesture Recognition (FG 2018)}, 2018, pp. 59--66.

\bibitem{Ekman70}
P.~Ekman, ``{Universal Facial Expressions of Emotion},'' \emph{California
  Mental Health}, vol. 8 (4), pp. 151--158, 1970.

\bibitem{Ruiz18}
A.~Ruiz-Garcia, N.~Webb, V.~Palade, M.~Eastwood, and M.~Elshaw, ``Deep learning
  for real time facial expression recognition in social robots,'' in
  \emph{Neural Information Processing}, L.~Cheng, A.~C.~S. Leung, and S.~Ozawa,
  Eds.\hskip 1em plus 0.5em minus 0.4em\relax Cham: Springer International
  Publishing, 2018, pp. 392--402.

\bibitem{Webb20}
N.~Webb, A.~Ruiz-Garcia, M.~Elshaw, and V.~Palade, ``Emotion recognition from
  face images in an unconstrained environment for usage on social robots,'' in
  \emph{2020 International Joint Conference on Neural Networks (IJCNN)}.\hskip
  1em plus 0.5em minus 0.4em\relax IEEE, 2020, pp. 1--8.

\bibitem{Han21}
H.~Han, O.~Karadeniz, E.~B. Sönmez, T.~Dalyan, and B.~Sarıoğlu, ``Facial
  expression recognition in the wild with application in robotics,'' in
  \emph{2021 6th International Conference on Computer Science and Engineering
  (UBMK)}, 2021, pp. 565--569.

\bibitem{Arnau22}
E.~Arnaud, A.~Dapogny, and K.~Bailly, ``Thin: Throwable information networks
  and application for facial expression recognition in the wild,'' \emph{IEEE
  Transactions on Affective Computing}, 2022.

\bibitem{Ramis20}
S.~Ramis, J.~M. Buades, and F.~J. Perales, ``Using a social robot to evaluate
  facial expressions in the wild,'' \emph{Sensors}, vol.~20, no.~23, p. 6716,
  2020.

\bibitem{Molla17}
A.~Mollahosseini, B.~Hasani, and M.~H. Mahoor, ``Affectnet: A database for
  facial expression, valence, and arousal computing in the wild,'' \emph{IEEE
  Transactions on Affective Computing}, vol.~10, no.~1, pp. 18--31, 2017.

\bibitem{Aoki22}
M.~Aoki, F.~Rea, D.~Jirak, G.~Sandini, T.~Yanagi, A.~Takamatsu, S.~Bouet, and
  T.~Yamamura, ``On the influence of social robots in cognitive tasks,''
  \emph{International Journal of Humanoid Robotics}, 2022, (accepted for
  publication).

\bibitem{Jirak22}
D.~Jirak, M.~Aoki, T.~Yanagi, A.~Takamatsu, S.~Bouet, T.~Yamamura, G.~Sandini,
  and F.~Rea, ``Is it me or the robot? a critical evaluation of human affective
  state recognition in a cognitive task,'' \emph{Frontiers in Neurorobotics},
  vol.~16, 2022.

\bibitem{Barro22}
P.~Barros and A.~Sciutti, ``Ciao! a contrastive adaptation mechanism for
  non-universal facial expression recognition,'' \emph{arXiv preprint
  arXiv:2208.07221}, 2022.

\end{thebibliography}

\end{document}